\documentclass{article}

\usepackage{PRIMEarxiv}

\usepackage[utf8]{inputenc} 
\usepackage[T1]{fontenc}    
\usepackage{hyperref}       
\usepackage{url}            
\usepackage{booktabs}       
\usepackage{amsfonts}       
\usepackage{nicefrac}       
\usepackage{microtype}      
\usepackage{lipsum}
\usepackage{graphicx}
\graphicspath{{media/}}     

\usepackage{graphicx}
\usepackage{amsmath}
\usepackage{amssymb}
\usepackage{booktabs}
\usepackage{xcolor}
\usepackage{subcaption}

\title{Similarity over Factuality: Are we making progress on multimodal out-of-context misinformation detection?
}

\usepackage{authblk}

\author[1, 2]{\small Stefanos-Iordanis Papadopoulos\thanks{Corresponding author} \ }
\author[1]{\small Christos Koutlis}
\author[1]{\small Symeon Papadopoulos}
\author[2]{\small Panagiotis C. Petrantonakis}

\affil[1]{\footnotesize Information Technology Institute, Centre for Research \& Technology, Hellas.}
\affil[2]{\footnotesize Department of Electrical \& Computer Engineering, Aristotle University of Thessaloniki.}
\affil[ ]{\textit {\{stefpapad,ckoutlis,papadop\}@iti.gr, \textit{ppetrant@ece.auth.gr}}}

\begin{document}
\maketitle

\begin{abstract}
Out-of-context (OOC) misinformation poses a significant challenge in multimodal fact-checking, where images are paired with texts that misrepresent their original context to support false narratives. 
Recent research in evidence-based OOC detection has seen a trend towards increasingly complex architectures, incorporating Transformers, foundation models, and large language models. 
In this study, we introduce a simple yet robust baseline, which assesses MUltimodal SimilaritiEs (MUSE), specifically the similarity between image-text pairs and external image and text evidence. 
Our results demonstrate that MUSE, when used with conventional classifiers like Decision Tree, Random Forest, and Multilayer Perceptron, can compete with and even surpass the state-of-the-art on the NewsCLIPpings and VERITE datasets. 
Furthermore, integrating MUSE in our proposed ``Attentive Intermediate Transformer Representations'' (AITR) significantly improved performance, by 3.3\% and 7.5\% on NewsCLIPpings and VERITE, respectively. 
Nevertheless, the success of MUSE, relying on surface-level patterns and shortcuts, without examining factuality and logical inconsistencies, raises critical questions about how we define the task, construct datasets, collect external evidence and overall, how we assess progress in the field. 
We release our code at: \url{https://github.com/stevejpapad/outcontext-misinfo-progress}.
\end{abstract}

\keywords{Multimodal Learning \and Deep Learning \and Misinformation Detection \and Automated Fact-checking}

\section{Introduction}
\label{sec:intro}

\begin{figure*}[t]
  \centering
   \includegraphics[width=1\linewidth]{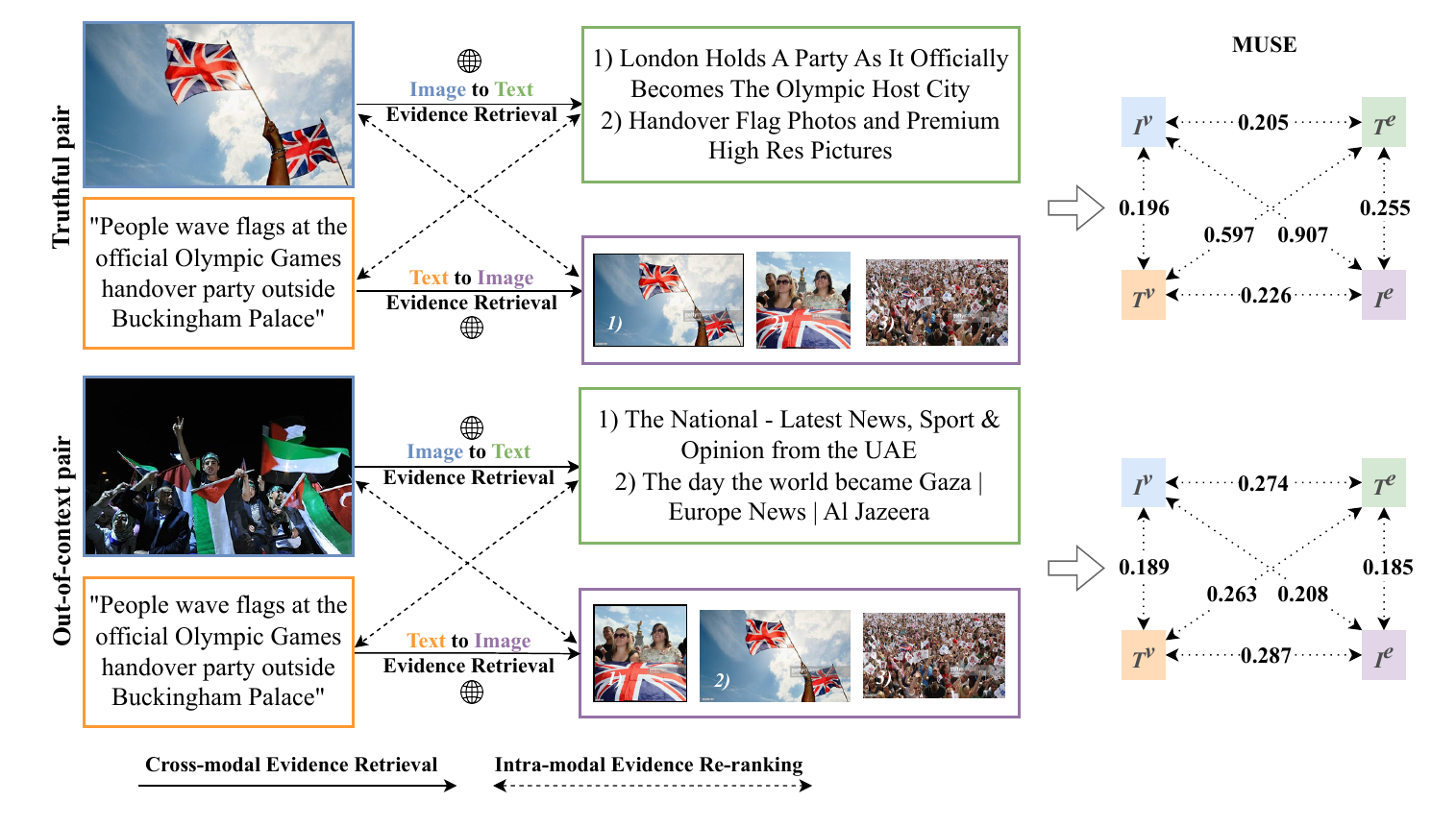}

   \caption{Samples from the NewsCLIPpings dataset, along with their retrieved and re-ranked evidence and their multimodal similarities.}
   \label{fig:example}
\end{figure*}

In recent decades, we have witnessed the proliferation of new types of misinformation, 
beyond fake news \cite{hirlekar2020natural} and manipulated images \cite{thakur2020recent}, including AI-generated ``DeepFakes'' \cite{masood2023deepfakes} and misinformation that spans multiple modalities such as images and texts\cite{wilson2023multimodal, comito2023multimodal}. 
In an effort to assist the work of human fact-checkers, researchers have been leveraging the power of deep learning
to automate certain aspects of the fact-checking process \cite{guo2022survey}, such as claim and stance detection, evidence and fact-check retrieval, and verdict prediction, among others\cite{kuccuk2020stance, yao2023end, vo2020facts, barron2020checkthat, pikuliak2023multilingual}. 

In this study, we focus on multimodal fact-checking, specifically targeting evidence-based out-of-context (OOC) detection, a topic that has recently gained significant attention from researchers. 
OOC misinformation involves the presentation of images with captions that distort or misrepresent their original context\cite{qian2023fighting}. 
Due to the lack of large-scale, annotated datasets for OOC detection, researchers have turned to algorithmic generation of OOC datasets \cite{aneja2023cosmos, luo2021newsclippings, papadopoulos2023synthetic} 
which have been used to train numerous  methods for OOC detection \cite{aneja2021mmsys, zhang2023detecting, mu2023self, dang2024overview, gu2024learning}, some of which leverage external information or evidence to further enhance detection accuracy \cite{abdelnabi2022open, zhang2023ecenet, qi2024sniffer, papadopoulos2023red}.
Overall, there is a trend towards increasingly complex architectures for OOC detection, including the integration of Transformers and memory networks \cite{abdelnabi2022open}, fine-tuning foundational vision-language models \cite{mu2023self}, incorporating modules for detecting relevant evidence \cite{papadopoulos2023red} and leveraging instruction tuning and large language models \cite{qi2024sniffer}, which generally translate to marginal improvements in performance. 

We develop a simple yet robust baseline that leverages MUultimodal SimilaritiEs (MUSE), specifically CLIP-based \cite{radford2021learning} similarities between image-text pairs under verification and across external image and text evidence. 
Our findings show that training machine learning classifiers, such as Decision Tree, Random Forest and Multi-layer Perceptron with MUSE can compete and even outperform much more complex architectures on NewsCLIPpings \cite{luo2021newsclippings, abdelnabi2022open} and VERITE \cite{papadopoulos2023red, papadopoulos2024verite} by up to 4.8\%. 
Furthermore, integrating MUSE within complex architectures, such as our proposed ``Attentive Intermediate Transformer Representations'' (AITR) can further improve performance, by 3.3\% on NewsCLIPpings and 7.5\% on VERITE, over the state-of-the-art (SotA). 

Nevertheless, our analysis reveals that the models primarily rely on shortcuts and heuristics based on surface-level patterns rather than identifying logical or factual inconsistencies. 
For instance, as illustrated in Fig.\ref{fig:example}, given a Truthful image-text pair,
we use the text to retrieve image evidence and the image to retrieve text evidence from the web. 
Due to the popularity of the NewsCLIPpings' sources (USA Today, The Washington Post, BBC, and The Guardian) search engines often retrieve the exact same or highly related images and texts as those under verification, which, after re-ranking, are selected as the likely evidence.  
This results in a high `image-to-evidence image' (0.907) `text-to-evidence text' (0.597) similarities.
In contrast, given the OOC image, we retrieve unrelated text evidence, leading to significantly lower similarity scores.
Consequently, a model can learn to rely on simple heuristics such as, if the image-text pair exhibits significant similarity both internally and with the retrieved (and re-ranked) evidence, then the pair is likely truthful; otherwise, it is OOC. 
Furthermore, we show that these models yield high performance only within a limited definition of OOC misinformation, where legitimate images are paired with otherwise truthful texts from different contexts. 
In contrast, their performance deteriorates when dealing with `miscaptioned images,' where images are de-contextualised by introducing falsehoods in their captions i.e. by altering named entities such as people, dates, or locations. 

These findings raise critical questions about how realistic and robust the current frameworks are, how we define the task, create datasets, collect external evidence, and, more broadly, how we assess progress in OOC detection and multimodal fact-checking.
In summary, we recommend future research to: 1) avoid training and evaluating methods solely on algorithmically created OOC datasets; 2) incorporate annotated evaluation benchmarks; 3) broaden the definition of OOC to include miscaptioned images, named entity manipulations, and other types of de-contextualization, and 4) to expand training datasets and collect external evidence accordingly.

\section{Related Work}
\label{sec:related}

Out-Of-Context (OOC) detection, also known as image re-purposing, multimodal mismatching, or ``CheapFakes'', involves pairing legitimate, non-manipulated, images with texts that misrepresent their context. 
Due to the lack of manually annotated and large-scale datasets, 
initial attempts to model out-of-context misinformation relied on randomly re-sampling image-text pairs \cite{aneja2023cosmos, jaiswal2017multimedia}, while more sophisticated methods now rely on hard negative sampling, creating out-of-context pairs that maintain semantic similarity
\cite{luo2021newsclippings, papadopoulos2023synthetic, biamby2022twitter}. 
In turn, multiple methods have been proposed for OOC detection that cross-examine and attempt to identify inconsistencies within the image-text pair without leveraging external evidence \cite{ aneja2021mmsys, zhang2023detecting, mu2023self, dang2024overview, gu2024learning}.
Another strand of research has focused on constructing multimodal misinformation datasets through weak annotations \cite{nakamura2020fakeddit} or named entity manipulations \cite{papadopoulos2023synthetic, sabir2018deep, muller2020multimodal} but, to the best of our knowledge, such datasets have not yet been enhanced with external evidence and used for multimodal fact-checking.
Nevertheless, professional fact-checkers\footnote{\url{https://www.factcheck.org/our-process}} rarely rely solely on internal inconsistencies between modalities and instead collect relevant external information, or evidence, that support or refute the claim under verification  \cite{guo2022survey, alhindi2018your}. 
Furthermore, prior studies on evidence-based OOC detection demonstrate significant performance improvements when leveraging external information \cite{abdelnabi2022open, qi2024sniffer, papadopoulos2023red}.

Specifically, Abdelnabi et al. \cite{abdelnabi2022open} enhanced the NewsCLIPpings dataset \cite{luo2021newsclippings} by collecting external evidence (See Section \ref{sec:dataset}) and developed the Consistency Checking Network (\textbf{CCN}) which examines image-to-image and text-to-text consistency using attention-based memory networks, that employ ResNet152 for images and BERT for texts, as well as a fine-tuned CLIP (ViT B/32) for additional multimodal features.
The Stance Extraction Network (\textbf{SEN}) employs the same encoders as CCN but enhances performance by semantically clustering external evidence to determine their stance toward the claim. 
It also integrates the co-occurrence of named entities between the text and textual evidence \cite{yuan2023support}. 
The Explainable and Context-Enhanced Network (\textbf{ECENet}) combines a coarse- and fine-grained attention network leveraging ResNet50, BERT and CLIP ViT-B/32 for multimodal feature extraction along with textual and visual entities \cite{zhang2023ecenet}. 
\textbf{SNIFFER} examines the ``internal consistency'' of image-text pairs and their ``external consistency'' with evidence with the use of a large language model, InstructBLIP, that is first fine-tuned for news captioning and then for OOC detection, utilizing GPT-4 to generate instructions that primarily focus on named entities while the Google Entity Detection API is used for extracting visual entities \cite{qi2024sniffer}.
Finally, the Relevant Evidence Detection Directed Transformer (\textbf{RED-DOT}) utilizes evidence re-ranking, element-wise modality fusion, guided attention and a Transformer encoder optimized with multi-task learning to predict the weakly annotated relevance of retrieved evidence \cite{papadopoulos2023red}.

On the whole, there is a noticeable trend toward increasing architectural complexity which typically translates into limited improvements in performance. 
In this study, we show how simple machine learning approaches can compete and even surpass complex SotA methods by simply leveraging multimodal 
similarities, which raises critical questions on how we define the task, collect data, external evidence and how we access progress in the field.

\begin{figure*}
  \centering
  \begin{subfigure}{0.49\linewidth}
   \includegraphics[width=1\linewidth]{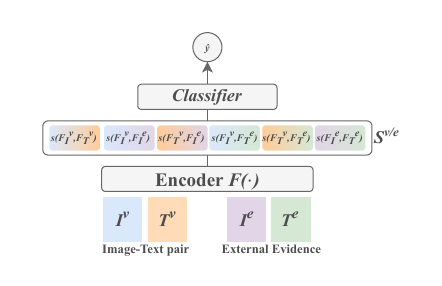}
    \caption{}
    \label{fig:muse}
  \end{subfigure}
  \hfill
  \begin{subfigure}{0.49\linewidth}
   \includegraphics[width=1\linewidth]{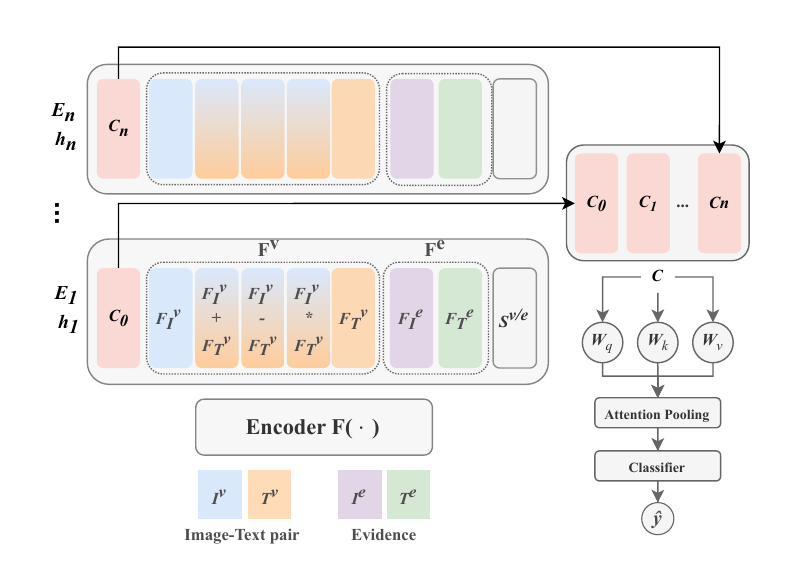}
    \caption{}
    \label{fig:aitr}
  \end{subfigure}
  \caption{Overview of the (a) MUSE and (b) AITR architectures.}
\end{figure*}

\section{Methodology}
\label{sec:method}

\subsection{Problem Formulation}
\label{sec:problem}
We define the task of evidence-based out-of-context detection as follows: given dataset $(I^{v}_i, T^{v}_i, I^{e}_i, T^{e}_i, y_i)_{i=1}^{N}$
where 
$I^{v}_i, T^{v}_i$
represent the image-text pair under verification, $I^{e}_i, T^{e}_i$ image and textual external information, or evidence,  retrieved for the pair and $y_i \in \{0,1\}$ is the pair's ground-truth label, being either \textit{truthful} (0) or \textit{out-of-context} (1), the objective is to train classifier $f: (\mathcal{I}^v, \mathcal{T}^v, \mathcal{I}^e, \mathcal{T}^e) \rightarrow \hat{y}^v$.

\subsection{Multimodal Similarities}

As shown in Fig.\ref{fig:muse}, given feature extractor $F(\cdot)$ and extracted features $F_{I^v}$, $F_{T^v}$, $F_{I^e}$, $F_{T^e}$, 
we use cosine similarity $s$ to calculate the Multimodal Similarities (MUSE) vector $S^{v/e}$ between $s(F_{I^{v}}, F_{T^{v}})$ image text pairs,
$s(F_{I^{v}}, F_{I^{e}})$ image to image evidence, 
$s(F_{T^{v}}, F_{I^{e}})$ text to image evidence, 
$s(F_{I^{v}}, F_{T^{e}})$ image to text evidence, 
$s(F_{T^{v}}, F_{T^{e}})$ text to text evidence and $s(F_{I^{e}}, F_{T^{e}})$ image evidence to text evidence. 

Afterwards, the $S^{v/e}$ vectors are used to train a machine learning classification such as Decision Tree (DT), Random Forest (RF) and Multi-layer Perceptron (MLP), denoted as MUSE-DT/RF/MLP, respectively, or are integrated within the ``Attentive Intermediate Transformer Representations'' (AITR) network. 

\subsection{Attentive Intermediate Transformer Representations}

Attentive Intermediate Transformer Representations (AITR) attempts to model how human fact-checkers may iterate multiple times over the claim and collected evidence during verification,  drawing various inferences and interpretations at each pass, exploring both general and fine-grained aspects and finally reassessing the entire process while assigning different weights to different aspects at each stage of analysis. 

As shown in Fig.\ref{fig:aitr}, AITR utilizes a stack of $n$ Transformer encoder layers $E(\cdot) = [E_1, E_2, \cdots, E_n]$ with $h = [h_1, h_2, \cdots, h_n]$ number of multi-head attention enabling 
both stable attention (e.g., $h = [8,8,8,8]$)
and granular attention, 
ranging from general to fine-grained (e.g., $h = [1,2,4,8]$) or from fine-grained to general (e.g., $h = [8,4,2,1]$). 
Given initial input:
\begin{equation}
x_0 = [C_0;F^v;F^e;S^{v/e}]
\end{equation}
where $C_0$ is a learnable classification token, 
$F^v$ represents element-wise modality fusion \cite{papadopoulos2023red} defined as 
$F^v = [F_{I^v};F_{T^v};F_{I^v}+F_{T^v};F_{I^v}-F_{T^v};F_{I^v}*F_{T^v}]$, 
$F_e = [F_{I^{e}}; F_{T^{e}}]$ and ``$;$'' denoting concatenation, intermediate Transformer outputs are given by:
\begin{equation}
x_{i} = E_i(x_{i-1}) \quad \text{for} \quad i \in \{1, 2, \ldots, n\}
\end{equation}
From each intermediate output, we extract the processed classification tokens 
$\mathcal{C} = [C_1, C_2, \ldots, C_n]$ 
and apply the scaled-dot product self-attention mechanism:
\begin{equation}
\mathcal{C}_a = \text{softmax}\left(\frac{Q \cdot K^T}{\sqrt{d}}\right) \cdot V
\end{equation}
with fully connected layers 
$Q = \textbf{W}_q \cdot \mathcal{C}$, 
$K = \textbf{W}_k \cdot \mathcal{C}$, 
$V = \textbf{W}_v \cdot \mathcal{C}$
and $\textbf{W}_q, \textbf{W}_k, \textbf{W}_v \in \mathbb{R}^{d \times d}$. 
Afterwards, we use average pooling to calculate $\mathcal{C}_p = \frac{1}{n} \sum_{i=1}^{n} \mathcal{C}_a[:, i, :]$ and a final classification layer to predict
$
\hat{y}^{v}=\textbf{W}_1\cdot \text{GELU}(\textbf{W}_0\cdot \mathcal{C}_p)
$
with $\textbf{W}_0\in\mathbb{R}^{1 \times d}$ and 
$\textbf{W}_1\in\mathbb{R}^{d \times 1}$. 

\section{Experimental Setup}

\subsection{Datasets}
\label{sec:dataset}
We utilize the NewsCLIPpings Merged/Balanced dataset, comprising 85,360 samples in total \cite{luo2021newsclippings}, 42,680 ``Pristine'' or truthful $\mathcal{I}^{v}, \mathcal{T}^{v}$ pairs sourced from credible news sources -as provided by the VisualNews dataset \cite{liu2021visual}- and 42,680 algorithmically created OOC pairs.
Specifically, OOC pairs are generated by mismatching the initial image or text with another, utilizing semantic similarities, either CLIP text-to-image or text-to-text similarities, SBERT-WK for text-to-text person mismatching, and ResNet Place for scene mismatching.
Furthermore, we utilize the VERITE evaluation benchmark \cite{papadopoulos2024verite} comprising 1,000 annotated samples, 338 truthful pairs, 338 miscaptioned images and 324 out-of-context pairs. 

\subsection{External Evidence}
For NewsCLIPpings, we use the external evidence $\mathcal{I}^{e}, \mathcal{T}^{e}$ as provided by \cite{abdelnabi2022open} comprising up to 19 text evidence and up to 10 image evidence for each $I^{v}_i, T^{v}_i$ pair, collected via Google API; totaling to 146,032 and 736,731 textual and image evidence, respectively. 
Specifically, the authors employ cross-modal retrieval, namely the text $T^{v}_i$ is used to retrieve potentially relevant image evidence $I^{e}_i$ and image $I^{v}_i$ to retrieve potentially relevant textual evidence $T^{e}_i$. 
We use the same Training, Validation and Testing sets as prior works to ensure comparability. 
For VERITE, we employ the external evidence as provided by \cite{papadopoulos2023red}. 

Instead of utilizing all provided evidence as in \cite{abdelnabi2022open}, we follow \cite{papadopoulos2023red}, in re-ranking the external evidence based on CLIP \cite{radford2021learning} intra-modal similarities (image-to-image evidence, text-to-text evidence). We only select the top-1 items, as leveraging additional items was shown to degrade performance by introducing less relevant and noisy information into the detection model. 

\subsection{Backbone Encoder}
Following \cite{abdelnabi2022open, papadopoulos2023red} we use the pre-trained 
CLIP ViT B/32 and ViT L/14  \cite{radford2021learning} 
as the backbone encoders in order to extract visual 
$F_{I^v}, F_{I^e}\in{R}^{d\times 1}$ and textual features $F_{T^v}, F_{T^e}\in{R}^{d\times 1}$ with dimensionality $d=512$ or $d=768$ for CLIP ViT B/32 and L/14, respectively. 
Unless stated otherwise, we employ L/14 while using B/32 only for comparability purposes with some older works. 
We use the ``\textit{openai}'' version of the models as provided by OpenCLIP\footnote{\url{https://github.com/mlfoundations/open_clip}}.

\subsection{Evaluation Protocol} 

We train each model on the NewsCLIPpings train set, tune the models' hyper-parameters on the validation set and report the best version's accuracy on the NewsCLIPpings test set and unless stated otherwise, 
as in Table \ref{tab:miscaptioned}, we report the ``True vs OOC'' accuracy for VERITE. 

To ensure comparability with \cite{papadopoulos2023red} on VERITE, 
we report the mean ``out-of-distribution cross-validation" (OOD-CV) accuracy for VERITE in Table \ref{tab:sota}. 
Specifically, we validate and checkpoint a model on a single VERITE fold (k=3) while evaluating its performance on the other folds. 
We then retrieve the model version (hyper-parameter combination) that achieved the highest mean validation score and report its mean performance of the testing folds. 

\subsection{Implementation Details}

We train AITR for a maximum of 50 epochs, with early stopping and check-pointing set at 10 epochs to prevent overfitting. 
The AdamW optimizer is utilized with $\epsilon=1e-8$ and $\text{weight decay}=0.01$. 
We employ a batch size of 512 and a transformer dropout rate of 0.1
During hyperparameter tuning, we explore learning rates $lr \in \{1e-4, 5e-5\}$, 
transformer feed-forward layer dimension $z\in\{256, 1024, 2048\}$ and for $h$ we try the following values $[4,4,4,4], [8, 8, 8, 8]$, $[1,2,4,8]$, $[8,4,2,1]$. 
In the ablation experiments that do not leverage intermediate transformer representations, we exclude the $h= [1,2,4,8]$ and $[8,4,2,1]$ configurations. 
To ensure reproducibility of our experiments, we use a constant random seed of 0 for PyTorch, Python Random, and NumPy.

\section{Experimental Results}

\subsection{Ablation and Comparative Studies}
\label{sec:ablation_comparative}

Table \ref{tab:aitr_ablation} presents the ablation study results for AITR which consistently achieves the highest performance among all ablation configurations, underscoring the importance of each component. 
Specifically, substituting the attention mechanism with max pooling or weighted pooling leads to a notable reduction in performance across both datasets. 
Similarly, using the default transformer encoder (Pooling = None) without leveraging intermediate representations lowers performance. 
Notably, the most critical component of AITR appears to be MUSE, as removing it significantly deteriorates the model's performance, especially on VERITE, in both AITR and the default Transformer encoder.
The best AITR performance was achieved with $h=[1, 2, 4, 8]$, $z=2048$ and learning rate $5e-5$.

\begin{table}
  \centering
  {{
  \begin{tabular}{llcc}
    \toprule
    Pooling & MUSE & NewsCLIPpings & VERITE \\
    \midrule
    None & No &  88.28 & 69.98 \\
    None & Yes &  93.19 & 78.13 \\
    Attention & No &  89.39 & 71.04 \\
    Max & Yes &  92.94 & 78.88 \\
    Weighted & Yes &  93.03 & 78.73 \\
    Attention & Yes &  93.31 & 81.00 \\
    \bottomrule
  \end{tabular}
  }}
  \caption{Ablation experiments of AITR using CLIP L/14 features.}
  \label{tab:aitr_ablation}
\end{table}

\begin{table}
  \centering
  {{
  \begin{tabular}{lcc}
    \toprule
    Method & NewsCLIPpings & VERITE \\
    \midrule
    CCN \cite{abdelnabi2022open} & 84.7 & - \\
    SEN \cite{yuan2023support} & 87.1 & - \\
    ECENet \cite{zhang2023ecenet} & 87.7 & - \\
    SNIFFER \cite{qi2024sniffer} & 88.4 & - \\
    RED-DOT (B/32) \cite{papadopoulos2023red} & 87.8 & 73.9 (0.5) \\
    RED-DOT (L/14) \cite{papadopoulos2023red} & 90.3 & 76.9 (5.4) \\
    \midrule
    MUSE-MLP (B/32) & 85.4 & 74.8 (3.8) \\
    MUSE-MLP (L/14) & 90.0 & 80.6 (4.1) \\
    AITR (B/32) & 89.8 & 76.5 (2.4) \\ 
    AITR (L/14) & \textbf{93.3} & \textbf{82.7} (6.1) \\

    \bottomrule
  \end{tabular}
  }}
  \caption{Comparative analysis of evidence-based approaches for OOC detection. 
  We report the ``True vs OOC'' accuracy and standard deviation (in parentheses) on VERITE under the OOD-CV protocol. 
  }
  \label{tab:sota}
\end{table}

In comparison with the current SotA, as shown in Table \ref{tab:sota}, MUSE-MLP competes and even outperforms much more complex architectures on NewsCLIPpings. Specifically, MUSE-MLP (90\%) performs similar to RED-DOT (90.3\%) while surpassing SNIFFER (88.4\%), ECENet (87.7\%), SEN (87.1\%) and CCN (84.7\%). 
Notably, MUSE-MLP also significantly outperforms RED-DOT on VERITE, with +4.8\% relative improvement. 
Furthermore, integrating MUSE within AITR, significantly outperforms the SotA on NewsCLIPpings by +3.3\% and VERITE by +7.5\%. 

While this study primarily focuses on evidence-based approaches, we may also note that MUSE-MLP with $s(F_{I^v}, F_{T^v})$ and no external evidence, achieves 80.7\% on NewsCLIPpings, as seen in Table \ref{tab:muse_ablation}, and thus outperforms complex and resource-intensive architectures such as the Self-Supervised Distilled Learning \cite{mu2023self} (71\%) that uses a fully fine-tuned CLIP ResNet50 backbone on NewsCLIPpings, the Detector Transformer \cite{papadopoulos2023synthetic} (77.1\%) and even competes against RED-DOT without evidence (81.7\%) \cite{papadopoulos2023red}. 

\subsection{Similarity Importance}
Furthermore, we examine the contribution of each similarity measure within $S^{v/e}$. 
Table \ref{tab:ml_baselines} illustrates the performance and feature importance by the Decision Tree and Random Forest classifiers. 
We observe that both classifier put the highest emphasis on the image-text pair similarity $s(F_{T^v}, F_{I^e})$ followed by image to image evidence $s(F_{I^v}, F_{I^e})$ and text to text evidence $s(F_{T^v}, F_{T^e})$. 

Table \ref{tab:muse_ablation} demonstrates an ablation of MUSE-MLP classifier on NewsCLIPpings (N) and VERITE (V), while excluding certain similarities. 
We observe that employing $S^{v/e}$ with all 6 similarity measures consistently achieves the highest overall accuracy (N=89.86, V=80.54) on both datasets. 
Therefore, each similarity measure contributes to some extend to the overall performance. 
Nevertheless, among single similarity experiments, we observe that 
$s(F_{T^v}, F_{I^e})$ and 
$s(F_{I^v}, F_{T^e})$ 
yield near-random performance
while image-text pair similarities $s(F_{I^v}, F_{T^v})$ 
yield the highest performance (N=80.69, V=70.89), followed by image to image evidence 
$s(F_{I^v}, F_{I^e})$ (N=79.86, V=68.02)
and then text to text evidence 
$s(F_{T^v}, F_{T^e})$ (N=71.83, V=52.19), where performance, especially on VERITE, drops significantly. 
Similarly, removing the image-text pair $s(F_{I^v}, F_{T^v})$ results in a notable drop performance with N=85.64\% and V=69.68\%.
Again, similar to the Random Forest and Decision Tree classifiers, it is 
$s(F_{I^v}, F_{T^v})$ that has the highest contribution and is followed by $s(F_{I^v}, F_{I^e})$ and $s(F_{T^v}, F_{T^e})$. 

\begin{table}
  \centering
  {{
  \begin{tabular}{lcc}
    \toprule
    & Random Forest & Decision Tree \\
    \midrule

    $s(F_{I^{v}}, F_{T^{v}})$ & 0.3424 & 0.5918\\ 
    $s(F_{I^{v}}, F_{I^{e}})$ & 0.2426 & 0.2149 \\
    $s(F_{T^{v}}, F_{I^{e}})$ & 0.0871 & 0.0178 \\
    $s(F_{I^{v}}, F_{T^{e}})$ & 0.0700 & 0.0547 \\
    $s(F_{T^{v}}, F_{T^{e}})$ &  0.1807 & 0.1154 \\
    $s(F_{I^{e}}, F_{T^{e}})$ & 0.0773 & 0.0054 \\

    \midrule

    NewsCLIPpings & 90.16 & 88.44 \\
    VERITE & 79.93 & 77.37 \\

    \bottomrule
  \end{tabular}
  }}
  \caption{Feature importance and performance by MUSE-RF and MUSE-DT. 
  }
  \label{tab:ml_baselines}
\end{table}

\begin{table*}
  \centering
  \footnotesize
  {\small{
  \begin{tabular}{cccccccc}
    \toprule
    $s(F_{I^{v}}, F_{T^{v}})$ & 
    $s(F_{I^{v}}, F_{I^{e}})$ & 
    $s(F_{T^{v}}, F_{I^{e}})$ &
    $s(F_{I^{v}}, F_{T^{e}})$ & 
    $s(F_{T^{v}}, F_{T^{e}})$ & 
    $s(F_{I^{e}}, F_{T^{e}})$ & 
    NewsCLIPpings & VERITE 
     \\
    
    \midrule

    \checkmark & -& -& -& -& -&  80.69 & 70.89 
    \\

    -& \checkmark & -& -& -& -&  79.86 & 68.02 
    \\
    
    -& -& \checkmark & -& -& -&   55.52 & 55.81 
    \\        
    
    -& -& -& \checkmark & -& -&  54.01 & 50.68 
    \\

    -& -& -& -& \checkmark & -&  71.83 & 52.19 
    \\
    
    -& -& -& -& -& \checkmark &   68.72 & 56.11 
    \\  
    
    -& \checkmark & -& -& \checkmark & -& 84.33 & 69.38 

 \\  

    \checkmark & -& -& -& \checkmark &  
     - & 84.99	& 74.06 
     \\

    \checkmark & \checkmark& -& -& - & -& 85.39 & 77.07 
    \\

    \checkmark & \checkmark & -& -& \checkmark &  -& 
      88.12 & 77.22 
      \\      
    
    \checkmark & \checkmark & -& -& \checkmark &  \checkmark 
      & 88.04 & 78.43 
      \\   

    \checkmark & \checkmark & \checkmark & \checkmark & - & \checkmark & 88.31 & 77.98 \\      

    \checkmark & - & \checkmark & \checkmark & \checkmark & \checkmark & 
    88.46 & 76.62 \\     
    
     -& \checkmark & \checkmark & \checkmark & \checkmark & \checkmark & 
      85.64 & 69.68 
      \\      

    \midrule

    \checkmark & \checkmark & \checkmark & \checkmark & \checkmark & \checkmark & 
     \textbf{89.96} & \textbf{80.54} 
     \\      
        
    \bottomrule
  \end{tabular}
  }}
  \caption{Ablation of MUSE-MLP. 
  }
  \label{tab:muse_ablation}
\end{table*}

\subsection{Performance with Limited Data}
\begin{figure}[t]
  \centering
   \includegraphics[width=0.8\linewidth]{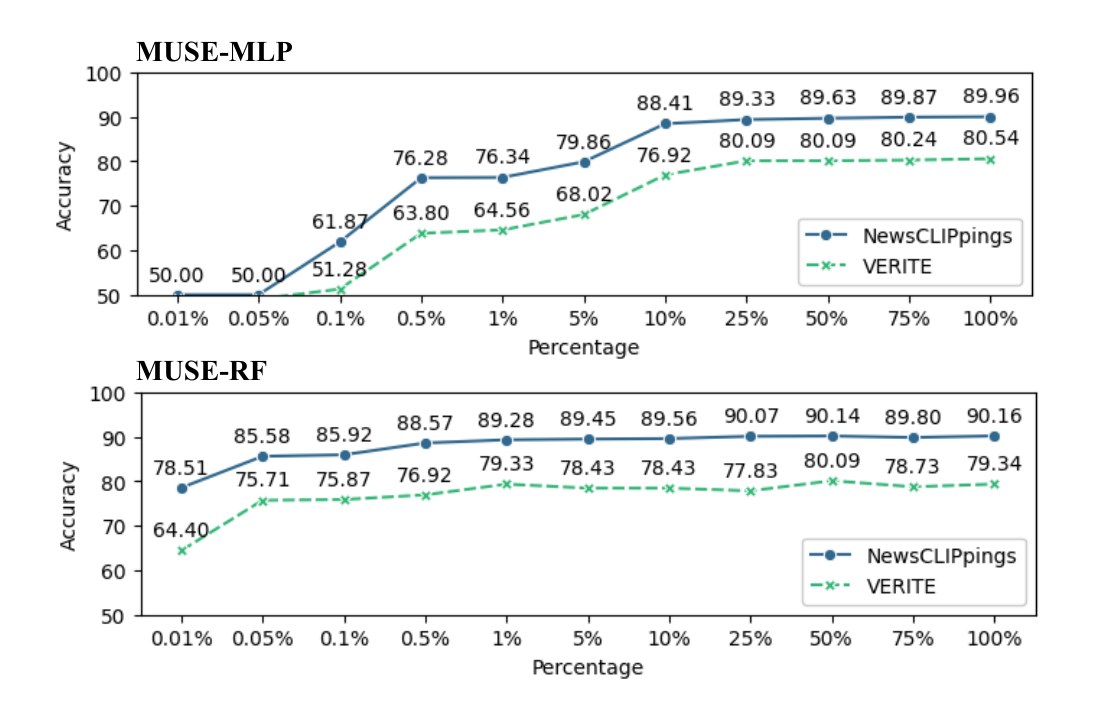}
   \caption{Performance of MUSE-MLP and MUSE-RF with limited training data. }
   \label{fig:percentage}
\end{figure}

As shown in Fig. \ref{fig:percentage}, 
MUSE-MLP maintains high performance on both dataset with only using 25\% of the NewsCLIPpings training set.
Notably, MUSE-RF maintains high performance even when trained with 1\% of the training set, which translates to only 710 samples. 
Surprisingly, even with 0.1\% and 0.05\% of the dataset, or 71 and 36 samples, respectively, the performance of MUSE-RF does not completely deteriorate. 
This means that the patterns that MUSE-RF relies on are simple enough that can be learned from even from a few tens or hundreds of samples.

\subsection{Pattern Analysis}
\begin{figure*}[t]
  \centering
   \includegraphics[width=1\linewidth]{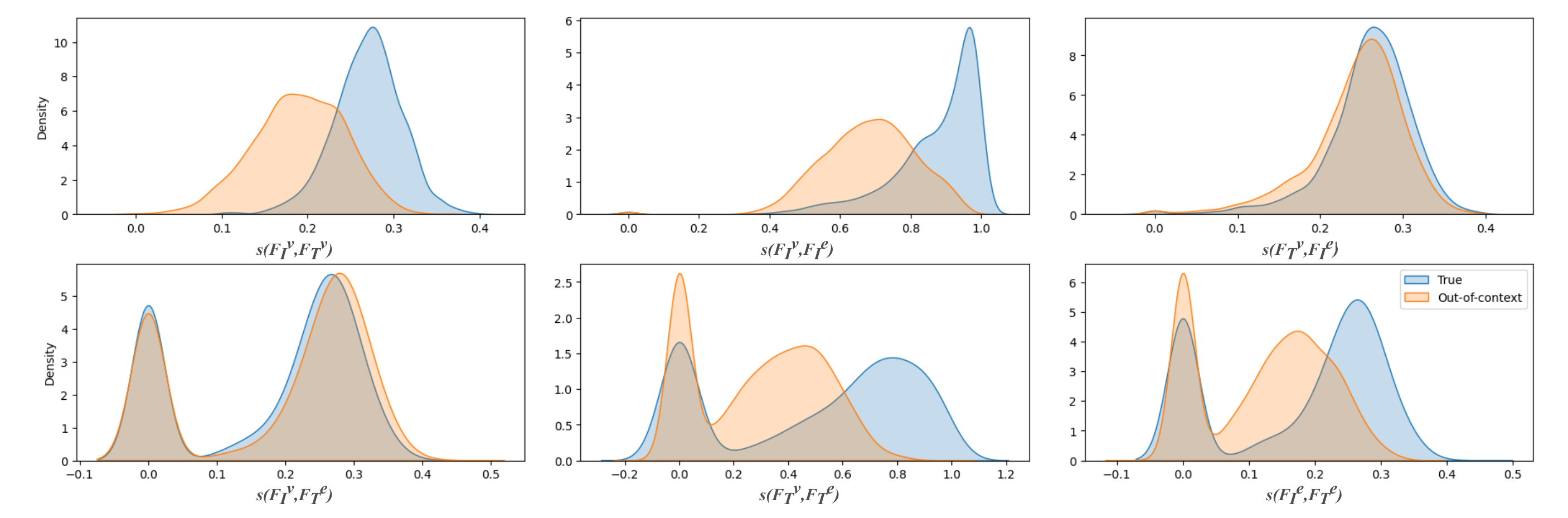}

   \caption{Distributions of similarity measures on NewsCLIPpings True and OOC classes. }
   \label{fig:distribution}
\end{figure*}

By examining Fig.\ref{fig:distribution}, illustrating the distributions of the 6 similarity measures in NewsCLIPpings, we observe clear differences between Truthful and OOC distributions, primarily on 
$s(F_{I^v}, F_{T^v})$, $s(F_{I^v}, F_{I^e})$, $s(F_{T^v}, F_{T^e})$ and $s(F_{I^e}, F_{T^e})$.
Indicatively, the median values of
$s(F_{I^v}, F_{T^v})$, 
$s(F_{I^v}, F_{I^e})$, 
$s(F_{T^v}, F_{T^e})$
are 0.27, 0.91, 0.63 for Truthful pairs
and 0.19, 0.69, 0.32 for OOC pairs.
In contrast, $s(F_{T^v}, F_{I^e})$ and $s(F_{I^v}, F_{T^e})$ demonstrate mostly overlapping distributions between True and OOC classes which explains why they result in near-random performance in single-similarity experiments of Table \ref{tab:muse_ablation}.

\begin{figure*}[t]
  \centering
   \includegraphics[width=1\linewidth]{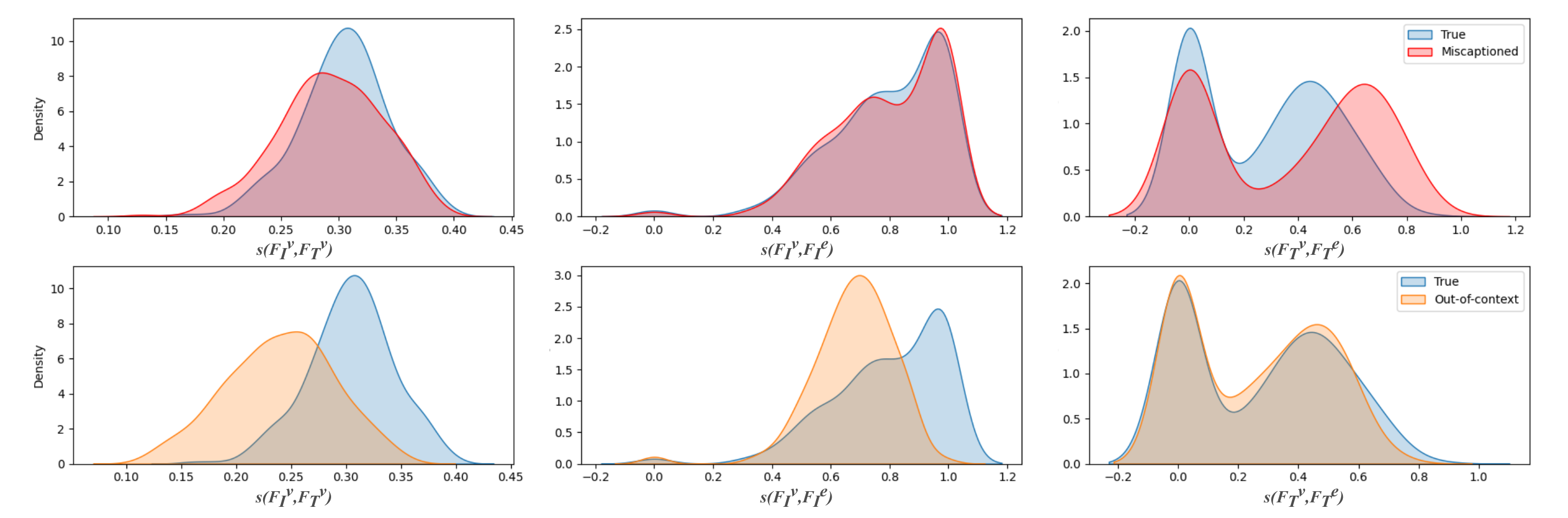}

   \caption{Distributions of similarity measures on VERITE, True vs OOC and True vs Miscaptioned classes. }
   \label{fig:verite_distribution}
\end{figure*}

In Fig. \ref{fig:verite_distribution} we observe that the similarity distributions of VERITE 
exhibits relatively similar ``True vs OOC'' distributions with NewsCLIPpings 
in terms of
$s(F_{I^v}, F_{T^v})$ and $s(F_{I^v}, F_{I^e})$ but not 
$s(F_{T^v}, F_{T^e})$ 
that have mostly overlapping distributions. 
Indicatively, the median values of
$s(F_{I^v}, F_{T^v})$, 
$s(F_{I^v}, F_{I^e})$, 
$s(F_{T^v}, F_{T^e})$
are 0.31, 0.83, 0.32 for Truthful pairs
and 0.24, 0.69, 0.28 for OOC pairs.

Importantly, we also observe that ``True vs Miscaptioned'' distributions are overlapping on $s(F_{I^v}, F_{I^e})$ and that
$s(F_{T^v}, F_{T^e})$ similarities of the ``Miscaptioned'' class is skewed towards higher similarity, 
with median values of 
0.29, 0.82 and 0.46 for $s(F_{I^v}, F_{T^v})$, 
$s(F_{I^v}, F_{I^e})$ and 
$s(F_{T^v}, F_{T^e})$, respectively, thus inverting the pattern found in NewsCLIPpings.
As a result, as seen in Table \ref{tab:miscaptioned}, while MUSE and AITR exhibit high performance on VERITE in terms of ``True vs OOC'', their performance completely degrades on the ``True vs Miscaptioned'' evaluation. 

\begin{table}
  \centering
  {{
  \begin{tabular}{lcc}
    \toprule
    & MUSE-MLP & AITR \\
    \midrule

    NewsCLIPpings & 89.96 & 93.31 \\
    \textcolor{gray}{True} 
    & \textcolor{gray}{92.21} 
    & \textcolor{gray}{93.34}	\\
    
    \textcolor{gray}{OOC} & \textcolor{gray}{87.72} & 
    \textcolor{gray}{93.28}
    \\

    \midrule

    VERITE ``True vs OOC'' & 80.54 & 81.00\\
    VERITE ``True vs Miscaptioned'' & 51.18 & 51.78 \\
    
    \textcolor{gray}{True} & \textcolor{gray}{89.94} & \textcolor{gray}{92.31} \\
    \textcolor{gray}{OOC} & \textcolor{gray}{70.77} & \textcolor{gray}{69.23} \\
    \textcolor{gray}{Miscaptioned} & \textcolor{gray}{12.43} & \textcolor{gray}{13.02} \\
    
    \bottomrule
  \end{tabular}
  }}
  \caption{Per-class performance of MUSE-MLP and AITR on NewsCLIPpings and VERITE.}
  \label{tab:miscaptioned}
\end{table}

\section{Discussion}
\label{sec:discussion}

Overall, the experimental results indicate that while our methods surpass the SotA, they primarily rely on shortcuts and simple heuristics rather than detecting logical and factual inconsistencies. 
This raises critical questions about the realism and robustness of the current OOC detection framework, as well as how we define the task, collect data and external information.

As discussed in Section \ref{sec:ablation_comparative}, our proposed methods, MUSE and AITR, reach high accuracy scores on NewsCLIPpings, surpassing the SotA.
It is important to note that OOC samples in NewsCLIPpings are generated by misaligning the original, truthful image-text pairs with other semantically similar images or texts, based on similarities by CLIP, ResNet and S-BERT features. 
Consequently, the truthful pairs tend to exhibit relatively higher cross-modal similarity, while OOC pairs demonstrate lower similarity, as seen in Fig.\ref{fig:distribution}.  
By relying on this simple relation, MUSE-MLP achieved a high accuracy of 81\%, without incorporating any external information. 

Integrating multimodal similarities with external evidence increased the detection accuracy to 90-93\%. 
To understand this result, it is essential to consider the role of the evidence retrieval process.
Following \cite{abdelnabi2022open}, external evidence is gathered through cross-modal retrieval, where the image $I^v$ is used to retrieve text evidence $T^e$ 
and the text $T^v$ is used to retrieve 
image evidence $I^e$. 
Afterwards, we re-rank the retrieved items  based on intra-modal similarity, meaning image-to-image and text-to-text comparisons.

Considering that the original truthful image-text pairs in NewsCLIPpings are sourced from VisualNews, which in turn collected pairs from four meainstream sources —USA Today, The Washington Post, BBC, and The Guardian— during the evidence collection process, it is highly likely that the same source article, or a highly related one, is retrieved. These conditions contribute significantly to the high accuracy observed.
For instance, as illustrated in Fig.\ref{fig:example} and discussed in Section \ref{sec:intro}, the Truthful pair exhibits very high $s(F_{I^v}, F_{I^e})$ (0.907) and a relatively high $s(F_{T^v}, F_{T^e})$ (0.597) similarity scores with the retrieved evidence 
while the OOC sample exhibits significantly lower scores. 
Although relying on such heuristics leads to high performance on the NewsCLIPpings dataset, the performance on the annotated OOC samples of VERITE is more limited, particularly for the OOC class (70\%), which is the primary focus of this task.
In terms of the ``True vs OOC'' evaluation on VERITE, our methods consistently outperform the SotA, though they display lower accuracy compared to NewsCLIPpings, achieving scores around 80-82\%. Additionally, there is a notable imbalance with higher accuracy for the Truthful class (90-92\%) compared to the OOC class (70\%).
More importantly, as discussed in relation to Fig.\ref{fig:verite_distribution}, 
MUSE and AITR can not generalize to `Miscaptioned' samples of VERITE. 
This is because miscaptioned images, as defined by Snopes and Reuters, typically involve images and texts that are highly related, but with some key aspect being misrepresented in the text, such as a person, date, or event.
CLIP features and similarities do not capture the subtle linguistic differences necessary to detect such cases. 

Nevertheless, there is certainly room for further improving OOC detection. 
Firstly, we recommend that future research in this field not only utilize algorithmically generated misinformation (e.g., NewsCLIPpings) but also incorporate annotated evaluation benchmarks such as VERITE. 
Additionally, it is crucial to implement evaluation tests and analyses that demonstrate the models' reliance on factuality and its ability to detect logical inconsistencies, rather than merely exploiting shortcuts and simple heuristics.
Furthermore, we find the current working definition of OOC to be rather limiting, as it focuses solely on truthful texts combined with mismatched (out-of-context) images. This definition may not fully capture the complexity of real-world OOC misinformation, where the texts themselves often contain falsehoods\footnote{
``Miscaptioned: photographs and videos that are ``real'' (i.e., not the product, partially or wholly, of digital manipulation) but are nonetheless misleading because they are accompanied by explanatory material \textbf{that falsely describes their origin, context, and/or meaning}.''
\url{https://www.snopes.com/fact-check/rating/miscaptioned}}. 
We recommend that future research in the field of automated fact-checking and evidence-based OOC detection expand their methods and training datasets to also include `miscaptioned images,'\cite{papadopoulos2024verite} which encompass cases where an image is decontextualized but key aspects of the image, such as the person, date, or event, are misrepresented within the text. 
To this end, weakly annotated datasets such as Fakeddit 
\cite{nakamura2020fakeddit}
and algorithmically created datasets based on named-entity manipulations, such as MEIR, TamperedNews and CHASMA \cite{papadopoulos2023synthetic, papadopoulos2024verite, sabir2018deep, muller2020multimodal}, can prove useful if they are augmented with external evidence and combined with existing OOC datasets such as NewsCLIPpings. 
Finally, we recommend future researchers to consider the problem of ``leaked evidence'' while collecting external information from the web \cite{glockner2022missing, chrysidis2024credible}.

\section{Conclusions}

In this study, we adress the challenge of out-of-context (OOC) detection by leveraging multimodal similarities (MUSE) between image-text pairs and external image and text evidence. 
Our results indicate that MUSE, even when used with conventional machine learning classifiers, can compete against complex architectures and even outperform the SotA on the NewsCLIPpings and VERITE datasets. 
Furthermore, integrating MUSE within our proposed ``Attentive Intermediate Transformer Representations'' (AITR) yielded further improvements in performance. 
However, we discovered that these models predominantly rely on shortcuts and simple heuristics for OOC detection rather than assessing factuality. 
Additionally, we found that these models excel only under a narrow definition of OOC misinformation, but their performance deteriorates under other types of de-contextualization. 
These findings raise critical questions about the current direction of the field, including the definition of OOC misinformation, dataset construction, and evidence collection and we discuss potential future directions to address these challenges.

\section*{Acknowledgments}
This work is partially funded by the project ``vera.ai: VERification Assisted by Artificial Intelligence” under grant agreement no. 101070093.

\bibliographystyle{unsrt}  
\bibliography{manuscript}

\begin{thebibliography}{10}

\bibitem{hirlekar2020natural}
Vaishali~Vaibhav Hirlekar and Arun Kumar.
\newblock Natural language processing based online fake news detection challenges--a detailed review.
\newblock In {\em 2020 5th International Conference on Communication and Electronics Systems (ICCES)}, pages 748--754. IEEE, 2020.

\bibitem{thakur2020recent}
Rahul Thakur and Rajesh Rohilla.
\newblock Recent advances in digital image manipulation detection techniques: A brief review.
\newblock {\em Forensic science international}, 312:110311, 2020.

\bibitem{masood2023deepfakes}
Momina Masood, Mariam Nawaz, Khalid~Mahmood Malik, Ali Javed, Aun Irtaza, and Hafiz Malik.
\newblock Deepfakes generation and detection: State-of-the-art, open challenges, countermeasures, and way forward.
\newblock {\em Applied intelligence}, 53(4):3974--4026, 2023.

\bibitem{wilson2023multimodal}
Anna Wilson, Seb Wilkes, Yayoi Teramoto, and Scott Hale.
\newblock Multimodal analysis of disinformation and misinformation.
\newblock {\em Royal Society Open Science}, 10(12):230964, 2023.

\bibitem{comito2023multimodal}
Carmela Comito, Luciano Caroprese, and Ester Zumpano.
\newblock Multimodal fake news detection on social media: a survey of deep learning techniques.
\newblock {\em Social Network Analysis and Mining}, 13(1):101, 2023.

\bibitem{guo2022survey}
Zhijiang Guo, Michael Schlichtkrull, and Andreas Vlachos.
\newblock A survey on automated fact-checking.
\newblock {\em Transactions of the Association for Computational Linguistics}, 10:178--206, 2022.

\bibitem{kuccuk2020stance}
Dilek K{\"u}{\c{c}}{\"u}k and Fazli Can.
\newblock Stance detection: A survey.
\newblock {\em ACM Computing Surveys (CSUR)}, 53(1):1--37, 2020.

\bibitem{yao2023end}
Barry~Menglong Yao, Aditya Shah, Lichao Sun, Jin-Hee Cho, and Lifu Huang.
\newblock End-to-end multimodal fact-checking and explanation generation: A challenging dataset and models.
\newblock In {\em Proceedings of the 46th International ACM SIGIR Conference on Research and Development in Information Retrieval}, pages 2733--2743, 2023.

\bibitem{vo2020facts}
Nguyen Vo and Kyumin Lee.
\newblock Where are the facts? searching for fact-checked information to alleviate the spread of fake news.
\newblock In {\em Proceedings of the 2020 Conference on Empirical Methods in Natural Language Processing (EMNLP)}, pages 7717--7731, 2020.

\bibitem{barron2020checkthat}
Alberto Barr{\'o}n-Cedeno, Tamer Elsayed, Preslav Nakov, Giovanni Da~San~Martino, Maram Hasanain, Reem Suwaileh, and Fatima Haouari.
\newblock Checkthat! at clef 2020: Enabling the automatic identification and verification of claims in social media.
\newblock In {\em European Conference on Information Retrieval}, pages 499--507. Springer, 2020.

\bibitem{pikuliak2023multilingual}
Mat{\'u}{\v{s}} Pikuliak, Ivan Srba, Robert Moro, Timo Hromadka, Timotej Smole{\v{n}}, Martin Meli{\v{s}}ek, Ivan Vykopal, Jakub Simko, Juraj Podrou{\v{z}}ek, and M{\'a}ria Bielikov{\'a}.
\newblock Multilingual previously fact-checked claim retrieval.
\newblock In {\em Proceedings of the 2023 Conference on Empirical Methods in Natural Language Processing}, pages 16477--16500, 2023.

\bibitem{qian2023fighting}
Sijia Qian, Cuihua Shen, and Jingwen Zhang.
\newblock Fighting cheapfakes: using a digital media literacy intervention to motivate reverse search of out-of-context visual misinformation.
\newblock {\em Journal of Computer-Mediated Communication}, 28(1):zmac024, 2023.

\bibitem{aneja2023cosmos}
Shivangi Aneja, Chris Bregler, and Matthias Nie{\ss}ner.
\newblock Cosmos: catching out-of-context image misuse using self-supervised learning.
\newblock In {\em Proceedings of the AAAI conference on artificial intelligence}, volume~37, pages 14084--14092, 2023.

\bibitem{luo2021newsclippings}
Grace Luo, Trevor Darrell, and Anna Rohrbach.
\newblock Newsclippings: Automatic generation of out-of-context multimodal media.
\newblock In {\em Proceedings of the 2021 Conference on Empirical Methods in Natural Language Processing}, pages 6801--6817, 2021.

\bibitem{papadopoulos2023synthetic}
Stefanos-Iordanis Papadopoulos, Christos Koutlis, Symeon Papadopoulos, and Panagiotis Petrantonakis.
\newblock Synthetic misinformers: Generating and combating multimodal misinformation.
\newblock In {\em Proceedings of the 2nd ACM International Workshop on Multimedia AI against Disinformation}, pages 36--44, 2023.

\bibitem{aneja2021mmsys}
Shivangi Aneja, Cise Midoglu, Duc-Tien Dang-Nguyen, Michael~Alexander Riegler, Paal Halvorsen, Matthias Nie{\ss}ner, Balu Adsumilli, and Chris Bregler.
\newblock Mmsys' 21 grand challenge on detecting cheapfakes.
\newblock {\em arXiv preprint arXiv:2107.05297}, 2021.

\bibitem{zhang2023detecting}
Yizhou Zhang, Loc Trinh, Defu Cao, Zijun Cui, and Yan Liu.
\newblock Detecting out-of-context multimodal misinformation with interpretable neural-symbolic model.
\newblock {\em arXiv preprint arXiv:2304.07633}, 2023.

\bibitem{mu2023self}
Michael Mu, Sreyasee Das~Bhattacharjee, and Junsong Yuan.
\newblock Self-supervised distilled learning for multi-modal misinformation identification.
\newblock In {\em Proceedings of the IEEE/CVF Winter Conference on Applications of Computer Vision}, pages 2819--2828, 2023.

\bibitem{dang2024overview}
Duc-Tien Dang-Nguyen, Sohail~Ahmed Khan, Michael Riegler, P{\aa}l Halvorsen, Anh-Duy Tran, Minh-Son Dao, and Minh-Triet Tran.
\newblock Overview of the grand challenge on detecting cheapfakes at acm icmr 2024.
\newblock In {\em Proceedings of the 2024 International Conference on Multimedia Retrieval}, pages 1275--1281, 2024.

\bibitem{gu2024learning}
Yimeng Gu, Mengqi Zhang, Ignacio Castro, Shu Wu, and Gareth Tyson.
\newblock Learning domain-invariant features for out-of-context news detection.
\newblock {\em arXiv preprint arXiv:2406.07430}, 2024.

\bibitem{abdelnabi2022open}
Sahar Abdelnabi, Rakibul Hasan, and Mario Fritz.
\newblock Open-domain, content-based, multi-modal fact-checking of out-of-context images via online resources.
\newblock In {\em Proceedings of the IEEE/CVF conference on computer vision and pattern recognition}, pages 14940--14949, 2022.

\bibitem{zhang2023ecenet}
Fanrui Zhang, Jiawei Liu, Qiang Zhang, Esther Sun, Jingyi Xie, and Zheng-Jun Zha.
\newblock Ecenet: Explainable and context-enhanced network for muti-modal fact verification.
\newblock In {\em Proceedings of the 31st ACM International Conference on Multimedia}, pages 1231--1240, 2023.

\bibitem{qi2024sniffer}
Peng Qi, Zehong Yan, Wynne Hsu, and Mong~Li Lee.
\newblock Sniffer: Multimodal large language model for explainable out-of-context misinformation detection.
\newblock In {\em Proceedings of the IEEE/CVF Conference on Computer Vision and Pattern Recognition}, pages 13052--13062, 2024.

\bibitem{papadopoulos2023red}
Stefanos-Iordanis Papadopoulos, Christos Koutlis, Symeon Papadopoulos, and Panagiotis~C Petrantonakis.
\newblock Red-dot: Multimodal fact-checking via relevant evidence detection.
\newblock {\em arXiv preprint arXiv:2311.09939}, 2023.

\bibitem{radford2021learning}
Alec Radford, Jong~Wook Kim, Chris Hallacy, Aditya Ramesh, Gabriel Goh, Sandhini Agarwal, Girish Sastry, Amanda Askell, Pamela Mishkin, Jack Clark, et~al.
\newblock Learning transferable visual models from natural language supervision.
\newblock In {\em International conference on machine learning}, pages 8748--8763. PMLR, 2021.

\bibitem{papadopoulos2024verite}
Stefanos-Iordanis Papadopoulos, Christos Koutlis, Symeon Papadopoulos, and Panagiotis~C Petrantonakis.
\newblock Verite: a robust benchmark for multimodal misinformation detection accounting for unimodal bias.
\newblock {\em International Journal of Multimedia Information Retrieval}, 13(1):4, 2024.

\bibitem{jaiswal2017multimedia}
Ayush Jaiswal, Ekraam Sabir, Wael AbdAlmageed, and Premkumar Natarajan.
\newblock Multimedia semantic integrity assessment using joint embedding of images and text.
\newblock In {\em Proceedings of the 25th ACM international conference on Multimedia}, pages 1465--1471, 2017.

\bibitem{biamby2022twitter}
Giscard Biamby, Grace Luo, Trevor Darrell, and Anna Rohrbach.
\newblock Twitter-comms: Detecting climate, covid, and military multimodal misinformation.
\newblock In {\em Proceedings of the 2022 Conference of the North American Chapter of the Association for Computational Linguistics: Human Language Technologies}, pages 1530--1549, 2022.

\bibitem{nakamura2020fakeddit}
Kai Nakamura, Sharon Levy, and William~Yang Wang.
\newblock Fakeddit: A new multimodal benchmark dataset for fine-grained fake news detection.
\newblock In {\em Proceedings of the Twelfth Language Resources and Evaluation Conference}, pages 6149--6157, 2020.

\bibitem{sabir2018deep}
Ekraam Sabir, Wael AbdAlmageed, Yue Wu, and Prem Natarajan.
\newblock Deep multimodal image-repurposing detection.
\newblock In {\em Proceedings of the 26th ACM international conference on Multimedia}, pages 1337--1345, 2018.

\bibitem{muller2020multimodal}
Eric M{\"u}ller-Budack, Jonas Theiner, Sebastian Diering, Maximilian Idahl, and Ralph Ewerth.
\newblock Multimodal analytics for real-world news using measures of cross-modal entity consistency.
\newblock In {\em Proceedings of the 2020 International Conference on Multimedia Retrieval}, pages 16--25, 2020.

\bibitem{alhindi2018your}
Tariq Alhindi, Savvas Petridis, and Smaranda Muresan.
\newblock Where is your evidence: Improving fact-checking by justification modeling.
\newblock In {\em Proceedings of the first workshop on fact extraction and verification (FEVER)}, pages 85--90, 2018.

\bibitem{yuan2023support}
Xin Yuan, Jie Guo, Weidong Qiu, Zheng Huang, and Shujun Li.
\newblock Support or refute: Analyzing the stance of evidence to detect out-of-context mis-and disinformation.
\newblock In {\em Proceedings of the 2023 Conference on Empirical Methods in Natural Language Processing}, pages 4268--4280, 2023.

\bibitem{liu2021visual}
Fuxiao Liu, Yinghan Wang, Tianlu Wang, and Vicente Ordonez.
\newblock Visual news: Benchmark and challenges in news image captioning.
\newblock In {\em Proceedings of the 2021 Conference on Empirical Methods in Natural Language Processing}, pages 6761--6771, 2021.

\bibitem{glockner2022missing}
Max Glockner, Yufang Hou, and Iryna Gurevych.
\newblock Missing counter-evidence renders nlp fact-checking unrealistic for misinformation.
\newblock In {\em Proceedings of the 2022 Conference on Empirical Methods in Natural Language Processing}, pages 5916--5936, 2022.

\bibitem{chrysidis2024credible}
Zacharias Chrysidis, Stefanos-Iordanis Papadopoulos, Symeon Papadopoulos, and Panagiotis Petrantonakis.
\newblock Credible, unreliable or leaked?: Evidence verification for enhanced automated fact-checking.
\newblock In {\em Proceedings of the 3rd ACM International Workshop on Multimedia AI against Disinformation}, pages 73--81, 2024.

\end{thebibliography}

\end{document}